\documentclass[10pt,twocolumn,letterpaper]{article}

\usepackage{cvpr}              

\usepackage{subcaption}

\usepackage{multirow, booktabs}

\usepackage{orcidlink}

\usepackage{multirow}
\usepackage{color, colortbl}
\usepackage{subcaption}
\usepackage{tabu}
\usepackage{pifont}
\usepackage{makecell}
\usepackage{paralist}
\usepackage{threeparttable}
\usepackage{enumitem}
\usepackage{hhline}
\usepackage{scalerel,xparse}
\usepackage{wrapfig}

\usepackage{comment}

\definecolor{cvprblue}{rgb}{0.21,0.49,0.74}
\PassOptionsToPackage{pagebackref,breaklinks,colorlinks,allcolors=cvprblue}{hyperref}\usepackage{hyperref}

\newcommand{\name}{\textsc{Basket}}

\def\logo{\hspace{2pt}\makebox[15pt][l]{\raisebox{-0.5ex}{\includegraphics[height=15pt]{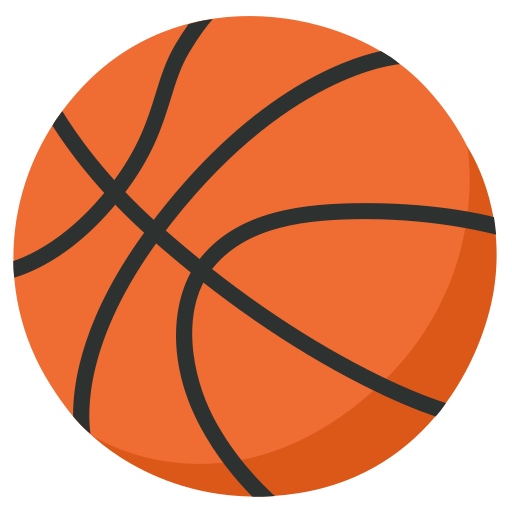}}}\hspace{3pt}}

\def\paperID{2295} 
\def\confName{CVPR}
\def\confYear{2025}

\title{\name \logo : A Large-Scale Video Dataset for Fine-Grained Skill Estimation}

\author{Yulu Pan\\
UNC Chapel Hill\\
{\tt\small yulupan@cs.unc.edu}
\and
Ce Zhang\\
UNC Chapel Hill\\
{\tt\small cezhang@cs.unc.edu}
\and
Gedas Bertasius\\
UNC Chapel Hill\\
{\tt\small gedas@cs.unc.edu}
\and
{\tt\small \url{https://sites.google.com/cs.unc.edu/basket}}
}

\begin{document}
\maketitle
\begin{abstract}

We present \name, a large-scale basketball video dataset for fine-grained skill estimation. \name~contains 4,477 hours of video capturing 32,232 basketball players from all over the world. Compared to prior skill estimation datasets, our dataset includes a massive number of skilled participants with unprecedented diversity in terms of gender, age, skill level, geographical location, etc. \name~includes 20 fine-grained basketball skills, challenging modern video recognition models to capture the intricate nuances of player skill through in-depth video analysis. Given a long highlight video (8-10 minutes) of a particular player, the model needs to predict the skill level (e.g., excellent, good, average, fair, poor) for each of the 20 basketball skills. Our empirical analysis reveals that the current state-of-the-art video models struggle with this task, significantly lagging behind the human baseline. We believe that \name~could be a useful resource for developing new video models with advanced long-range, fine-grained recognition capabilities. In addition, we hope that our dataset will be useful for domain-specific applications such as fair basketball scouting, personalized player development, and many others. Dataset and code are available at \url{https://github.com/yulupan00/BASKET}.
\end{abstract}
    
\section{Introduction}
\label{sec:intro}

From art painting with creativity to scoring a three-point shot with precision, we observe and admire human skills in many domains. People in various fields strive to master skills to continuously push the boundaries of our bodies, minds, and the world. Subsequently, others are drawn to watch these people demonstrate their exceptional skills for entertainment and also as a source of inspiration to improve themselves. For example, a sport like basketball, which requires many different skills, attracts 400 million fans worldwide. AI tools enabling a deeper comprehension of such fine-grained skills could lead to many practical applications, from the development of personalized coaching tools to enhancing a fan's watching experience with an expert-level commentary and skill analysis. 
\begin{table*}[!t]
    \centering
    \renewcommand{\arraystretch}{1.03}
    \small
    \begin{tabular}{llllll}
    \hline
    \textbf{Dataset} & \textbf{Total Video Hours} & \textbf{Num. Participants} & \textbf{Avg. Video Length (S)} & \textbf{Num. Skills} & \textbf{Num. Domains} \\ \hline
    JIGSAWS \cite{ahmidi2017dataset} & 3.5 & 8 & 120 & 3 & 1 \\
    BEST \cite{Doughty_2019_CVPR} & 26 & 400 & 188 & 5 & 5 \\
    MIT-Dive \cite{pirsiavash2014assessing} & 0.1 & 159 & 2.5 & 1 & 1 \\ 
    MIT-Skating \cite{pirsiavash2014assessing} & 7.3 & 150 & 175 & 1 & 1 \\ 
    UNLV-Dive \cite{parmar2017learning} & 0.4 & 370 & 3.8 & 1 & 1 \\ 
    MTL-AQA \cite{parmar2019and} & 1.5 & 1412 & 4.1 & 1 & 1 \\ 
    FineDiving \cite{xu2022finediving} & 3.5 & 3000 & 4.2 & 1 & 1 \\ 
    LOGO \cite{zhang2023logo} & 11 & 200 & 204 & 1 & 1 \\ 
    Fis-V \cite{xu2019learning} & 23.6 & 500 & 170 & 1 & 1 \\ 
    FP-Basket \cite{bertasius2017baller} & 10.3 & 48 & \textbf{780} & 1 & 1 \\
    EgoExoLearn \cite{huang2024egoexolearn} & 120 & 747 & 580 & 8 & \textbf{8 }\\
    Ego-Exo4D \cite{grauman2024ego} & 1286 & 740 & 156 & 8 & \textbf{8} \\
    \hline
    \rowcolor{blue!10} \textbf{\name} (Ours) & \textbf{4477} & \textbf{32232} & 500 & \textbf{20} & 1 \\ \hline
    \end{tabular}
    \vspace{0.1cm}
    \caption{Comparison with existing skill estimation video datasets.
    Our proposed \name~dataset significantly surpasses previous datasets in scale and diversity, with 4,477 video hours and 32,232 participants. Additionally, compared to most prior datasets, our dataset provides a larger number of fine-grained skills and includes videos with a longer average duration. 
    }
    \label{dataset_compare}
\end{table*}

\begin{figure}[t]
    \centering
    \includegraphics[width=1.0\columnwidth]{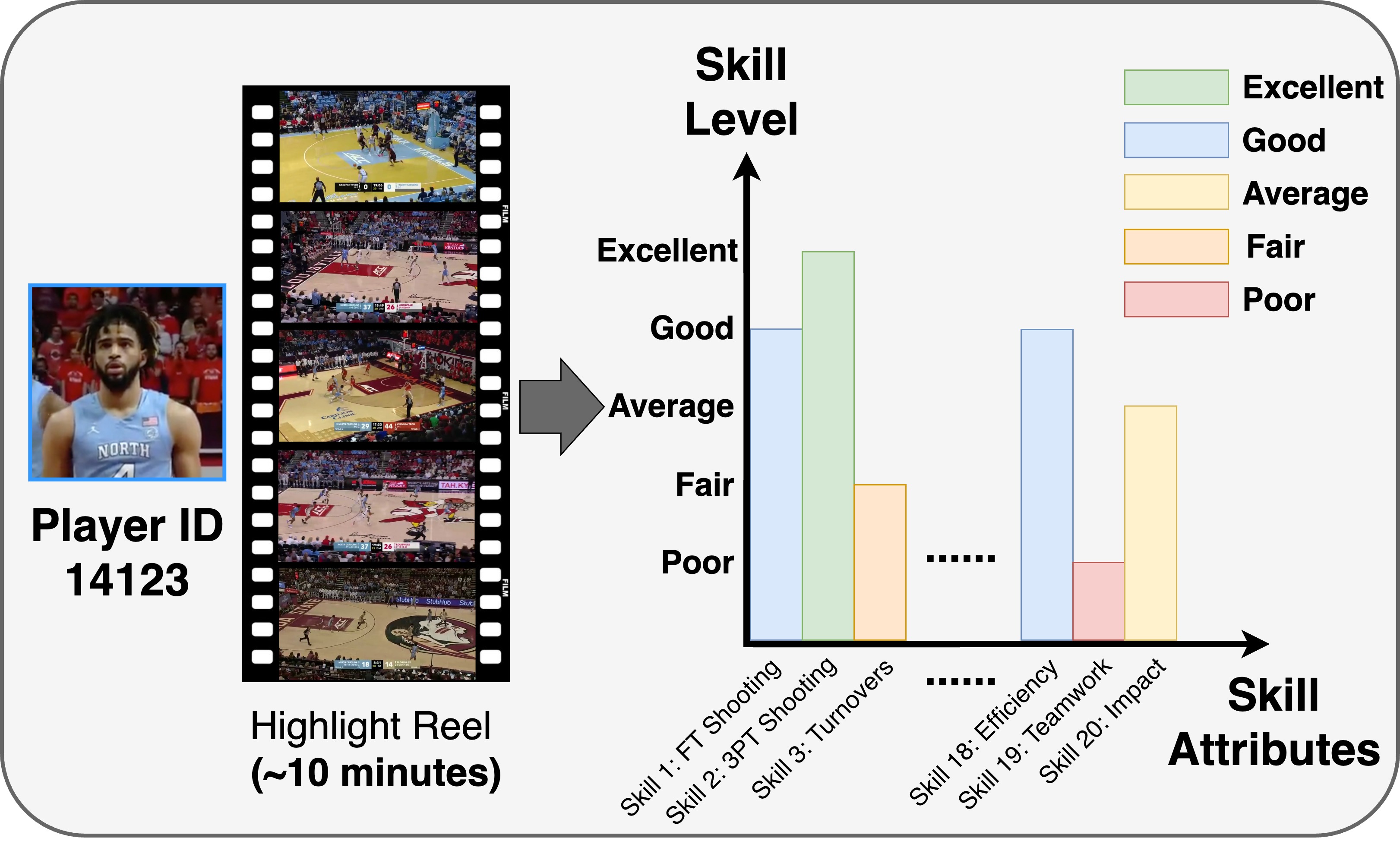}
    \caption{
    An illustration of our fine-grained skill estimation task. Given a long highlight video (8-10 minutes in length)  that captures many plays of a particular player, the model needs to predict the skill level for 20 fine-grained basketball skills (e.g., three-point shooting, rebounding, passing, etc.). Each skill is rated on a 5-level scale, from ``Poor" to ``Excellent."
    } 
    \label{img:flowchart}
    \vspace{-0.2cm}
\end{figure}

Recent years have witnessed remarkable progress in video models for recognizing human activities with greater precision and finer detail~\cite{bertasius2021space,wu2022memvit,li2023unmasked,wang2023videomae,ma2022x,wang2024internvideo2,li2024videomamba,hussein2019videograph,zhou2023adafocus}. The video recognition field has evolved from recognizing basic actions in short clips, like those in the Kinetics~\cite{carreira2017quo} dataset, to identifying more complex and subtle motions in longer contexts, such as those in EgoSchema~\cite{mangalam2023egoschema}. However, despite progress in general action recognition, the fine-grained skill estimation task, which requires recognizing how well the actions are performed, has received limited attention and is still highly underexplored. 

One of the main limiting factors in skill estimation is the lack of large-scale video training datasets, which have effectively fueled the progress in many other video recognition domains~\cite{carreira2017quo,bain2021frozen,miech2019howto100m}. 
As shown in Table~\ref{dataset_compare}, the existing skill estimation datasets~\cite{ahmidi2017dataset,xu2019learning,parmar2019and,xu2022finediving,zhang2023logo,shao2020finegym,pirsiavash2014assessing,grauman2024ego} are typically very small and lack significant participant diversity. Even the recently collected Ego-Exo4D~\cite{grauman2024ego} dataset, which took 2 years, 15 institutions, and millions of dollars, only contains 800 participants, which is insufficient to train modern, data-hungry video recognition models. As a result, most existing skill estimation models are brittle and have limited generalization capabilities. 

We introduce \name~(\textbf{Ba}sketball \textbf{Sk}ill \textbf{E}stimation over \textbf{T}ime), a large-scale basketball video dataset for advancing fine-grained skill estimation. Our dataset contains over 4,400 hours of video, capturing 32,232 basketball players from 21 basketball leagues worldwide. \name~also offers an unprecedented level of diversity in terms of participant characteristics (gender, race, age, nationality, skill level), geographic location (4 continents and over 30 countries), and the number of fine-grained skill attributes (e.g., three-point shooting, rebounding, passing, defending, etc.). We chose basketball as our primary domain for the following four reasons. First, basketball offers huge participant diversity and lots of video data that we can use to train skill estimation models. Second, basketball involves many fine-grained skills, making the skill estimation task more challenging and interesting. Third, in basketball, most players have similar visual appearances, necessitating the models to recognize fine-grained visual/skill cues rather than scene/background biases, as is common in traditional video recognition datasets. Lastly, the skilled basketball activities of each player are captured across many basketball games over several months, necessitating temporal video understanding, which many modern video modes struggle with. As shown in Table \ref{dataset_compare}, our dataset significantly exceeds prior skill estimation datasets in terms of scale and variety of skills represented. 

We formulate fine-grained skill estimation as a multi-way video classification problem. Specifically, given a long 8-10 minute highlight video that captures many plays of a particular player, we aim to estimate that player's skill level over 20 fine-grained skill attributes, such as three-point shooting, passing, rebounding, defending, and many others (discussed in Section~\ref{sec:basket}). Each skill is categorized into five levels: ``Excellent," ``Good," ``Average," ``Fair," and ``Poor." This is a challenging problem for several reasons. First, it requires video models to process long-form video inputs of 8-10 minutes, which is difficult for modern video recognition models. Second, due to the difficulty of obtaining bounding-box player annotations, the model has to jointly learn to identify the player of interest in each video and estimate that player's skill. Lastly, compared to traditional object or action recognition tasks, our skill estimation task focuses on a higher-level understanding of a person's skill, thus requiring fundamentally different visual representations that can capture subtle and nuanced skill cues rather than spatial or background cues needed for coarse action recognition~\cite{carreira2017quo,gu2018ava,soomro2012ucf101,kuehne2011hmdb}.

Our empirical experiments with various video recognition models, including Internvideo2~\cite{wang2024internvideo2}, VideoMamba~\cite{li2024videomamba}, X-CLIP~\cite{ma2022x}, and others, reveal that these models struggle to achieve good results on our \name~benchmark. Specifically, we report that the best performing VideoMamba~\cite{li2024videomamba} only achieves \textbf{28.50\%} accuracy (compared to 20\% random baseline performance). In contrast, the most experienced human annotators can achieve as high as \textbf{72\%} on this challenging task, highlighting the gap in fine-grained video recognition models' capabilities. We hope that our work will encourage the skill estimation community to use our large-scale video benchmark to develop new and more powerful fine-grained skill estimation models.

\begin{figure*}[!t]
    \centering
    \vspace{-0.3cm}
    \includegraphics[width=2\columnwidth]{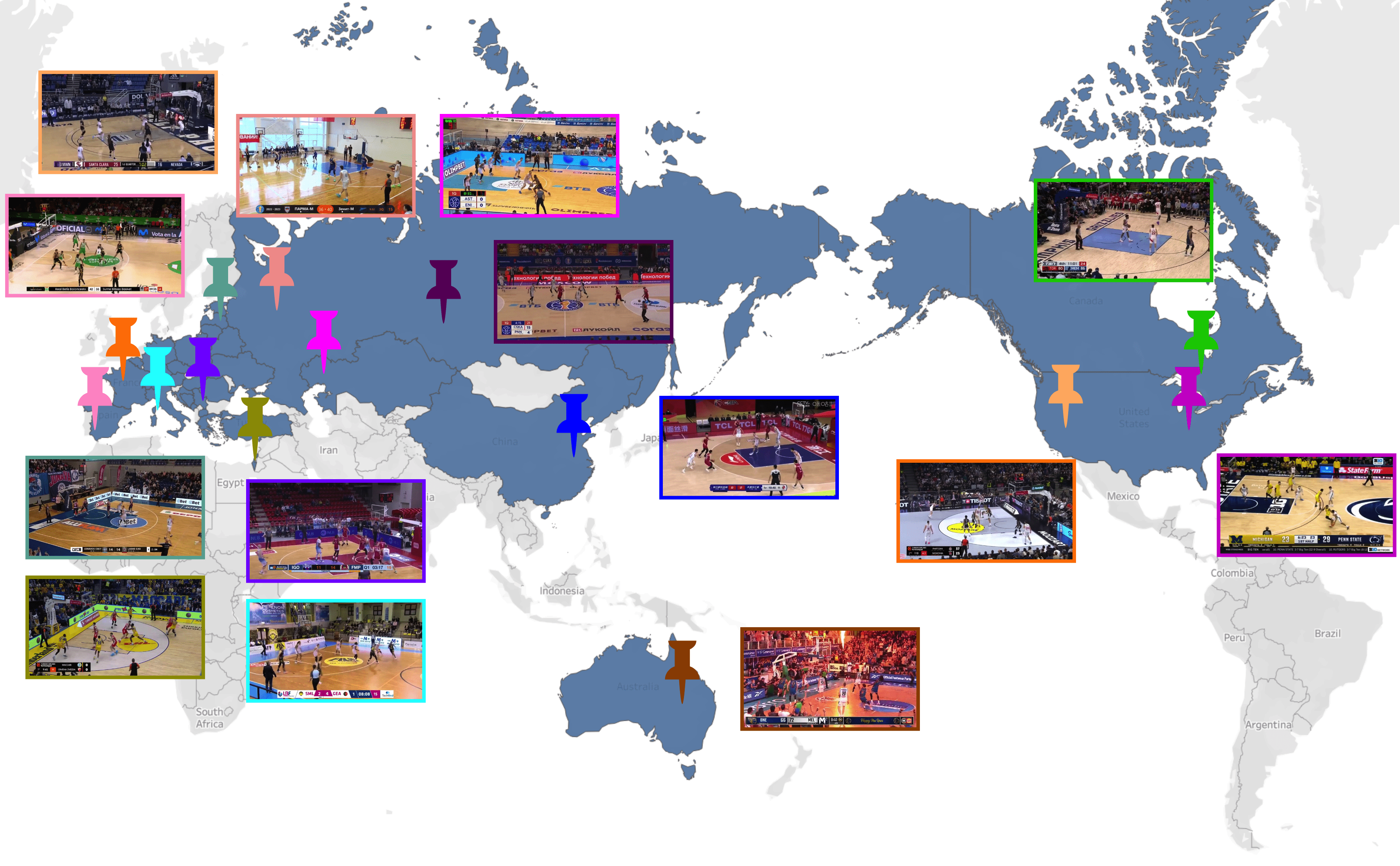}
    \caption{
    \name~is a large-scale video dataset containing 4,477 hours of video and capturing 32,232 basketball players from 21 basketball leagues worldwide. Here, we showcase geographic location diversity of our dataset, i.e., it captures basketball players from 4 continents and more than 30 countries. Each pin marks the approximate geographic location of the visualized basketball game (the color of a pin corresponding to the border of the visualized game).
    } 
    \label{img:geo}
    \vspace{-0.2cm}
\end{figure*}

\section{Related Work}
\label{sec:formatting}

\textbf{Skill Estimation Datasets.} Skill estimation has recently gained interest, particularly in domains where understanding and assessing human performance is essential. Such tasks and datasets are more challenging than general video recognition datasets such as Kinetics~\cite{carreira2017quo}, AVA~\cite{gu2018ava}, UCF-101~\cite{soomro2012ucf101}, and HMDB51~\cite{kuehne2011hmdb}.
In sports, most existing skill-related datasets focus on diving actions. MIT-Dive~\cite{pirsiavash2014assessing}, UNLV-Dive~\cite{parmar2017learning}, MTL-AQA~\cite{parmar2019and}, and FineDiving~\cite{xu2022finediving} are diving datasets with fine-grained annotations of action procedures, accompanied by the official judge score. 
LOGO~\cite{zhang2023logo} dataset consists of multi-person long-form artistic swimming competition videos annotated with human assessment scores. 
Other sports-based skill datasets are related to basketball~\cite{bertasius2017baller}, figure staking~\cite{pirsiavash2014assessing}, and golf~\cite{mcnally2019golfdb}. 
Beyond sports, several datasets have been collected to evaluate surgical and daily skills. JIGSAWS~\cite{ahmidi2017dataset} is a video dataset designed to evaluate the surgical skills of three procedures. BEST~\citep{Doughty_2019_CVPR} contains videos across five daily tasks and uses pairwise rankings for skill assessment tasks. EgoExoLearn~\cite{huang2024egoexolearn} includes egocentric and demonstration videos to estimate skill proficiency for daily and lab tasks. Most recently, Ego-Exo4D~\cite{grauman2024ego} is a multi-view egocentric and exocentric proficiency estimation benchmark covering 8 physical and procedural scenarios. 
Unlike these prior datasets, our \name~ dataset offers a much greater scale, number of skills, and participant diversity. 

\noindent\textbf{Skill Estimation Methods.} Several existing methods tackle action quality assessment and skill estimation tasks. 
The work in~\cite{ismail2018evaluating} proposed a CNN-based model to segment kinematic data into hierarchical features for surgical skill classification. Subsequently, a multi-path framework~\cite{Liu_2021_CVPR} was developed to combine various skill aspects from surgical videos. Building on this, the method in~\cite{Yu_2021_ICCV} proposed a contrastive regression framework integrated with group-aware regression to assess surgical proficiency.
For sports, MTL-AQA~\cite{parmar2019and} incorporates a multitask learning framework with spatiotemporal features extracted by 3D CNNs for diving quality assessment. FineDiving~\cite{xu2022finediving} proposes a procedure-aware action quality assessment approach that uses a temporal segmentation attention module to analyze spatial, semantic, and temporal correspondences in diving. 
LOGO~\cite{zhang2023logo} introduces a group-aware attention module that integrates spatial-temporal group dynamics to model relations between artistic swimmers in a scene. RICA$^{2}$~\cite{majeedi2024rica} uses a deep probabilistic model that integrates human score rubrics and models prediction uncertainty for action quality assessment in diving and surgical skills. 
NS-AQA~\cite{okamoto2024hierarchical} introduces a hierarchical neuro-symbolic approach for evaluating diving quality. 
Compared to these prior works, our main objective is not to propose a novel skill estimation method but to introduce a new, challenging, large-scale, fine-grained skill estimation dataset.

\noindent \textbf{Video Recognition Models.} Traditional video recognition models~\cite{bertasius2021space,li2023unmasked,wang2023videomae,li2024videomamba} are primarily designed to recognize coarse action classes in short video clips such as Kinetics~\cite{carreira2017quo}. More recent work explores fine-grained video action recognition, such as recognizing gymnastic actions~\cite{shao2020finegym,li2021multisports}, diving procedures~\cite{parmar2019and,xu2022finediving}, basketball moves~\cite{yan2020social,xu2024finesports}, and figure skating~\cite{liu2020fsd,liu2021temporal}. 
Furthermore, models such as Temporal Query Network~\cite{zhang2021temporal} and PoseConv3D~\cite{duan2022revisiting} have been proposed to address the challenges in detailed temporal video understanding. In addition to fine-grained video recognition, recent years have witnessed many methods for long video modeling, including MeMViT~\cite{wu2022memvit}, 
TimeCeption~\cite{hussein2019timeception},
ViS4mer~\cite{islam2022long},
VideoGraph~\cite{hussein2019videograph}, AdaptFocus~\cite{zhou2023adafocus}, and 
VideoMamba~\cite{li2024videomamba}. 
Finally, recent video models~\cite{ma2022x,wang2024internvideo2, xu2021videoclip, zhao2023learning, pramanick2023egovlpv2, lin2022egocentric} 
have focused on learning powerful video representations from multimodal video-text data. Compared to these prior video approaches, in this work, we introduce a new, challenging fine-grained skill estimation dataset and show that modern video recognition models lack the capabilities needed to excel at it.

\begin{figure}[!t]
    \centering
    \vspace{-0.3cm}
    \includegraphics[width=0.95\columnwidth]{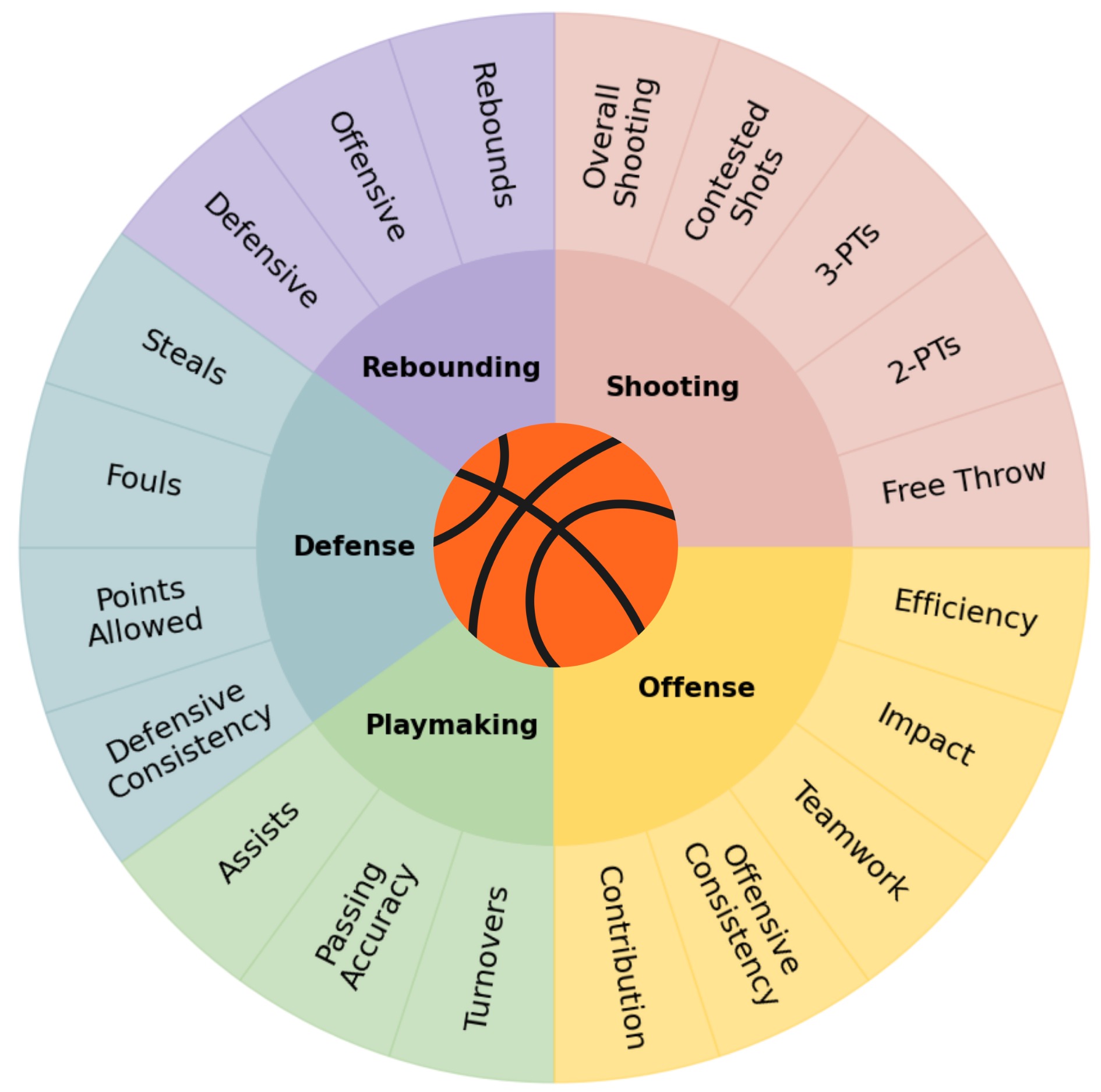}
    \vspace{-0.2cm}
    \caption{
    Our \name~dataset covers five coarse basketball skill categories and twenty fine-grained skills, focusing on the evaluation of multi-faceted skill understanding of basketball players. 
    } 
    \label{img:skill_breakdown}
    \vspace{-0.3cm}
\end{figure}
\section{The \name~Dataset}\label{sec:basket}

Here, we discuss the construction and characteristics of our new large-scale skill estimation video dataset, \name.

\subsection{Dataset Construction}

\textbf{Collecting Full-Game Videos.} We use a basketball game replay archive to collect basketball videos across many leagues worldwide. This results in roughly 66,000 basketball games. Each full-game video also contains a transcript annotated by experts with timestamped player-event instances (e.g., At 3:03 in the first quarter, Steph Curry makes a three-point shot). 

\noindent\textbf{Player Highlight Video Generation.} To generate player highlight videos, we first curate a list of 32,232 players from 21 unique basketball leagues worldwide. Since the players' skills can vary from year to year, we treat the same players in different seasons as different subjects. We use the aligned full-game video footage and timestamped player-event instances (described above) to extract basketball events associated with each player. Since the number of events per player can be large, we randomly select $50$ events for each player and then extract video clips around the selected events' timestamps. Each event clip is, on average, $10$ seconds long to capture the preceding context and the outcome of the player's actions. Finally, the selected clips are shuffled and combined into a single highlight video of approximately $9$ minutes in length. All videos are 704x400 in resolution and $30$ FPS.

\begin{figure*}[h]
    \centering
    \hspace*{3ex}
    \includegraphics[width=1.95\columnwidth]{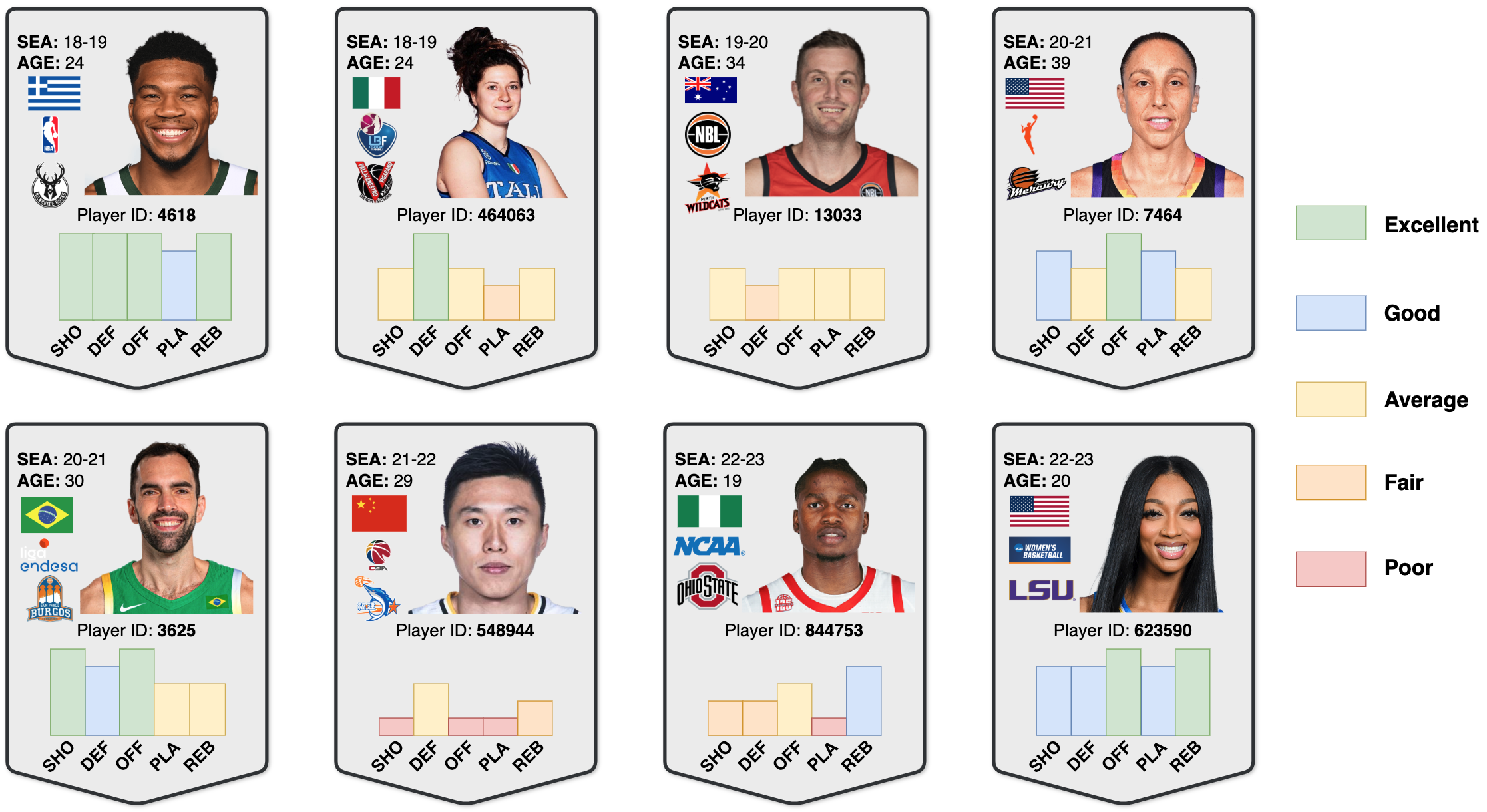}
    \vspace{-0.1cm}
    \caption{
    Visualizing some of the players from \name~ dataset. Our dataset offers unprecedented player diversity in terms of player nationality, age, gender, race, experience, and skill. The left side of each profile card displays the season, player nationality, league, and club. Skill levels are derived by averaging finer-level skills within each coarse category (as described in Section \ref{sec:basket}). \textbf{SHO}: Shooting, \textbf{DEF}: Defense, \textbf{OFF}: Offense, \textbf{PLA}: Playmaking, \textbf{REB}: Rebounding 
    } 
    \label{img:players}
    \vspace{-0.2cm}
\end{figure*}

\noindent\textbf{Player Skill Annotations.} We also obtain expert-annotated ranks of the players for $20$ fine-grained basketball skills, such as three-point shooting, passing, rebounding, and others (see Section~\ref{subsec:dataset_stats} for a more detailed analysis). The skill level labels are obtained by the basketball experts who review the game footage and annotate all the events in the game. These annotations are then aggregated and used to rank each player in 20 skill categories.
To divide all players into five skill level categories (i.e., ``Excellent," ``Good," ``Average," ``Fair," and ``Poor"), we sort the ranks of all players within the same league same season from highest to lowest and divide them evenly into 5 skill levels. We believe that five skill level categories make the task sufficiently challenging but still solvable. 
We also observe that different leagues (e.g., college vs. professional) may exhibit significant skill level differences. To account for these differences and avoid player comparisons across different leagues, we construct the skill labels separately for each league and season. During model training/inference, the model has to implicitly learn to predict skill levels specific to each basketball league.

\subsection{Dataset Statistics}\label{subsec:dataset_stats}
\label{sec:skill}

\name~consists of 32,232 player videos with an average duration of 500 seconds. 7,563 of these videos include women players. The dataset encompasses 21 basketball leagues from over 30 countries across 4 continents (See Figure~\ref{img:geo}) over 6 seasons, from 2017 to 2023. The dataset also represents players with a broad range of experience and age, from college-level athletes to professional players with 10-20 years of experience. Figure \ref{img:players} showcases some of the players from our \name~dataset, highlighting diversity in nationality, age, gender, race, experience, and skill. 

As shown in Figure~\ref{img:skill_breakdown}, \name~encompasses 20 fine-grained basketball skills. These skills can be divided into five broader categories, including shooting, playmaking, defense, rebounding, and offense. These coarse categories can be further divided into finer-grained skills (See Figure~\ref{img:skill_breakdown}). 
\section{Fine-Grained Skill Estimation Task}\label{sec:task}

Given a long player highlight video, we aim to classify each of the 20 basketball skills into 5 levels from ``Poor" to ``Excellent." Formally, given a video $V = \{x_t\}_{t=1}^{T} \in \mathbb{R}^{3 \times T \times H \times W}$ with $T$ frames, where each frame $x_t \in \mathbb{R}^{3 \times H \times W}$ represents an RGB image of size $H \times W$, the model needs to output a set of skill level predictions $Y = \{y_s\}_{s=1}^{S}$ where $S$ is the number of skills (i.e., 20 in our setting). Each prediction $y_s \in \mathbb{R}^L$ represents the probabilities for each skill level, and $L$ is the number of skill levels (i.e., 5 in our setting). 

This task presents several key challenges. First, the video model must be able to handle long video inputs. This includes long-term reasoning capabilities about different short-term segments within the video, as well as efficient long-video processing to enable scalable training on many long videos. Second, since the video inputs do not contain bounding box annotations for the player of interest, the model needs to jointly identify (implicitly) the recurring player (i.e., the player of interest) in all the video segments and then estimate that player's skills. This can be very challenging, as basketball videos often contain scenes with fast-moving motions, heavy occlusions of players, and significant camera cuts. Third, since each player's highlight is aggregated from only 50 event clips, the model needs to extrapolate that player's skill level by analyzing subtle cues of player performance, such as differences in techniques, player posture/pose, efficiency/effectiveness of their actions, decision-making, team dynamics, and many other factors. Achieving this requires fundamentally different perception capabilities focusing on nuanced skill cues rather than coarse background cues as is typical with standard action recognition tasks~\cite{carreira2017quo,gu2018ava,soomro2012ucf101,kuehne2011hmdb}.

\begin{table*}[!t] 
    \centering
    \setlength\tabcolsep{3pt}
    \renewcommand{\arraystretch}{1.15}
    \vspace{-0.3cm}
    \begin{tabular}{l c c c | c}
        \toprule
        \textbf{Method} & \textbf{Pretraining Datasets} & \textbf{\# Frames} & \textbf{\# Params (M)} & \textbf{Test Acc. (\%)} \\
        \hline
        Random Baseline & - & - & - & 20.00 \\
        SigLIP~\cite{zhai2023sigmoid} & WLI+K400 & 64 & 203 &  21.85 \\
        MeMViT~\cite{wu2022memvit} & K400 & 64 & 212 & 23.01 \\
        LLaVA-OneVision~\cite{li2024llava} & CC+BLIP+SD+UR+SG & 32 & 8200 & 23.84 \\
        X-CLIP~\cite{ma2022x} & WIT+K400 & 16 & 196 & 24.37 \\
        VideoMAE2~\cite{wang2023videomae} & UH+K400 & 32 & 1012 & 24.43 \\
        TimeSformer~\cite{bertasius2021space} &  IN21K+K400 & 96 & 121 & 25.21 \\
        UnmaskedTeacher~\cite{li2023unmasked} & K710+CC+VG+SBUC+CC15M+WV12M & 32 & 90 & 26.97 \\
        InternVideo2~\cite{wang2024internvideo2} & LA+WV10M+WV2M+SC+K710 & 32 & 1024 & 27.52 \\
        \rowcolor{gray!10} VideoMamba~\cite{li2024videomamba} & IN1K+K400 & 64 & 74 & \textbf{28.50} \\
        \bottomrule
    \end{tabular}%
    \caption{Comparison of various video recognition models on our fine-grained skill benchmark, \name. All experiments were conducted with a uniform video frame sampling strategy with a 224x224 spatial video resolution by fine-tuning each model with its best configuration. These results show that none of the methods achieves over 30\%  accuracy, indicating a large room for future improvement. \textbf{IN}: ImageNet, \textbf{K}: Kinetics, \textbf{WLI}; WebLI, \textbf{WIT}: WebImageText, \textbf{UH}: UnlabeledHybrid, \textbf{LA}: LAION, \textbf{WV}: WebVid, \textbf{SC}: Self-collected, \textbf{CC}: COCO, \textbf{VG}: Visual Genome, \textbf{SBUC}: SBU Captions, \textbf{UR}: UReader, \textbf{SD}: SynDOG, \textbf{SG}: ShareGPT4V
    }
    \label{main_result}
\end{table*}

\section{Experimental Setup}\label{sec:setup}

In the following section, we provide details on our dataset splits, evaluation metrics, and baselines we consider.

\noindent\textbf{Dataset Splits.} We split 32,232 players from \name~into training, validation, and test splits using the 7:1:2 ratio. Since our considered videos span seasons from 2017 to 2023, we ensure that each unique player is consistently included within the same set (i.e., all Stephen Curry's videos are in the training set). Additionally, we exclude videos from the 2017-2018 season from the training data and use them only for testing to validate the model generalization to previously unseen season data. Furthermore, we also exclude the videos from 4 of our selected leagues (from 21 total leagues) from the training data and use it only for testing to validate the generalization to the data from the previously unseen basketball leagues. In total, we use roughly 19,500 players for training, 2,800 for validation, 5,600 for testing, and 4,500 for generalization experiments. 

\noindent\textbf{Evaluation Metrics.} We use top-1 accuracy to evaluate the skill estimation performance in each skill category. To obtain a single skill estimation metric, we average the accuracies across all 20 fine-grained skill categories. 

\noindent\textbf{Baselines.} Since our benchmark and task do not have well-established baselines, we implement a number of our own baselines to measure the performance on our \name~benchmark. \underline{\textit{TimeSformer~\cite{bertasius2021space}}} is a convolution-free video classification model that relies solely on self-attention over space and time, adapting the Vision Transformer (ViT) for video understanding. \underline{\textit{MeMViT~\cite{wu2022memvit}}} introduces a memory-augmented multiscale vision transformer for long-term video recognition, processing videos incrementally and caching the extracted information into memory for referencing past context. \underline{\textit{VideoMAEV2~\cite{wang2023videomae}}} uses a masked video autoencoding strategy to learn powerful spatiotemporal features in a self-supervised manner. 
\underline{\textit{Unmasked Teacher (UMT)~\cite{li2023unmasked}}} learns strong video representations by aligning unmasked video tokens with the representations of foundational image models. \underline{\textit{InternVideo2~\cite{wang2024internvideo2}}} learns spatiotemporally consistent video representations via distillation from models like InternVL and VideoMAEv2, which capture spatial, temporal, and multi-modal information from the video. \underline{\textit{VideoMamba~\cite{li2024videomamba}}} adopts a bidirectional selective state-space model (SSM) architecture for scalable and memory-efficient long video processing. \underline{\textit{X-CLIP~\cite{ma2022x}}} extends CLIP to video-text retrieval with multi-grained contrastive learning, aligning coarse-grained and fine-grained visual features. \underline{\textit{SigLIP~\cite{zhai2023sigmoid}}} uses a pairwise sigmoid loss to learn visual representations from large-scale image-language data. To extend SigLIP to video, we apply temporal pooling on individually extracted frame-level features. 
\underline{\textit{LLaVA-OneVision~\cite{li2024llava}}} introduces a large vision-language model that achieves impressive cross-scenario generalization transfer across many image and video tasks. To adapt this model to our setting, we converted our dataset into text format and asked the model to output skill predictions for each category in the text format. 

\noindent\textbf{Implementation Details.} For all our experiments we use eight NVIDIA RTX A6000 GPUs, each with 48G of memory. All models are fine-tuned to optimal performance using the best available pre-trained checkpoints and hyperparameters. We ablate on several key hyperparameters in Section \ref{sec:results}. Each video model uses 20 classification heads (one linear layer each) for predicting each of the 20 skill categories. 
For all experiments, we sample the video frames uniformly and resize the shorter side of the frame to 224 pixels.

\section{Experimental Results}\label{sec:results}

In this section, we first present the results of all the baseline models on our \name~benchmark. Afterward, we present the human study results and the ablation studies conducted on the best-performing video model.

\begin{table*}[t]
    \centering
    \begin{minipage}[t]{0.36\linewidth}
        \centering
        \renewcommand{\arraystretch}{1.25}
        \setlength\tabcolsep{3pt} 
        {\fontsize{9}{9}\selectfont 
            \begin{tabular}{c | c c c c c}
                \textbf{Season} & 18-19 & 19-20 & 20-21 & 21-22 & 22-23\\
                \Xhline{1.0pt}
                \textbf{Acc. (\%)} & 27.12 & 27.41 & 27.36 & 29.38 & 29.92 \\
            \end{tabular}
        }
        \subcaption{Test Accuracy Across Seasons}
        \label{tab:ablation_season}
    \end{minipage}
    \hspace{0.1cm}
    \begin{minipage}[t]{0.22\linewidth}
        \centering
        \renewcommand{\arraystretch}{1.25}
        \setlength\tabcolsep{3pt}
        {\fontsize{9}{9}\selectfont 
            \begin{tabular}{c | c c}
                \textbf{Gender} & Male & Female\\
                \Xhline{1.0pt}
                \textbf{Acc. (\%)} & 27.51 & 31.33 \\
            \end{tabular}
        }
        \subcaption{Test Accuracy by Gender}
        \label{tab:ablation_gender}
    \end{minipage}
    \hspace{0.1cm}
    \begin{minipage}[t]{0.35\linewidth}
        \centering
        \renewcommand{\arraystretch}{1.25}
        \setlength\tabcolsep{3pt}
        {\fontsize{9}{9}\selectfont 
            \begin{tabular}{c | c c c c}
                \textbf{Location} & N. America & Europe & Asia & Australia\\
                \Xhline{1.0pt}
                \textbf{Acc. (\%)} & 29.68 & 25.58 & 24.24 & 23.00 \\
            \end{tabular}
        }
        \subcaption{Test Accuracy by Location}
        \label{tab:ablation_continent}
    \end{minipage}
    \vspace{-0.1cm}
    \caption{We study how VideoMamba generalizes to videos with different 1) seasons, 2) player gender, and 3) geographic locations.
    }
    \vspace{-0.2cm}
    \label{tab:ablation_breakdown}
\end{table*}

\subsection{Main Results}

In Table \ref{main_result}, we present the performance of the latest state-of-the-art video recognition models on our fine-grained skill estimation benchmark, \name. These results suggest that all models perform poorly, with none achieving more than 30\% accuracy (the random baseline being 20\%). From the results, we observe that the best-performing model is VideoMamba, likely due to its long-range temporal reasoning capability, which is crucial for capturing the extended context in our skill estimation task. Furthermore, we note that while models pretrained on large-scale image and video datasets, such as UnmaskedTeacher and InternVideo2, show moderate improvement over other less performant models, they remain low overall, underscoring that the task cannot be solved by simply scaling the image/video data from the Web. Lastly, our results show that the state-of-the-art vision-language models, such as LLaVA-OneVision and SigLIP, are among the worst performers. 
These low accuracies support our earlier claims that to do well on our challenging \name~benchmark, we need new models with fundamentally different visual recognition capabilities. 

\begin{figure}[t]
    \centering
    \includegraphics[width=0.95\columnwidth]{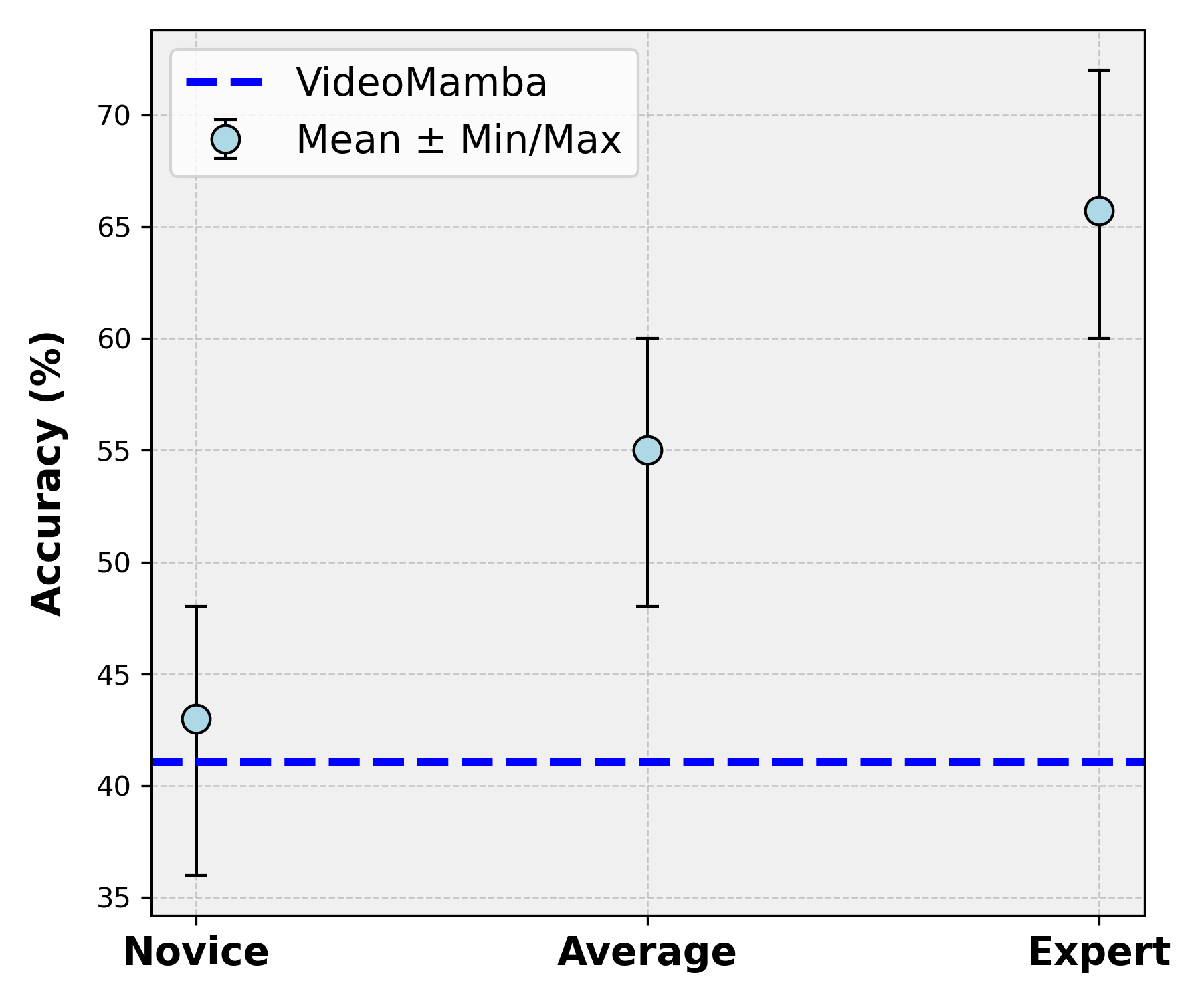}
    \caption{
    We visualize our human study results on \name, with subjects grouped by their expertise level (i.e., novice, average, expert). For each group, we visualize the mean accuracy and the min/max ranges. The blue dashed line indicates the performance of our best model, VideoMamba~\cite{li2024videomamba}. To ensure that the time needed to complete the study is reasonable, every subject is asked to watch videos of 5 uniformly selected players and classify 5 selected skills into 3 skill levels (i.e., ``Poor", ``Average," and ``Excellent"). Our VideoMamba baseline, which was not trained on these players, is also tested in this exact setting. Our results highlight the gap between model and human performance, especially for the human subjects with high expertise.
    } 
    \label{img:human_study}
    \vspace{-0.4cm}
\end{figure}

\subsection{Human Evaluation} 

Next, we conduct a study to assess human performance on our challenging skill estimation \name~benchmark. To ensure that the time needed to complete the study is reasonable, instead of using all 20 skill categories (see Figure~\ref{img:skill_breakdown}), we select one fine-grained skill from each broader skill category (i.e., shooting, rebounding, defense, playmaking, and offense) to assess 5 skills in total. Additionally, our initial human study experiments revealed that classifying each skill into 5 skill levels (i.e., ``Poor," ``Fair," ``Average," ``Good," and ``Excellent") takes too long for human subjects to complete the study (i.e., more than 1 hour for a single session). Therefore, our finalized human study involves asking each human subject to evaluate 5 players across 5 skills using 3 levels for each skill. 

\begin{table*}[t]
    \centering
    \renewcommand{\arraystretch}{1.25}
    \setlength\tabcolsep{8.0pt}
    {\fontsize{9}{9}\selectfont 
    
    \begin{minipage}[t]{0.275\linewidth}
        \vspace{0pt}
        \centering
        \resizebox{\textwidth}{!}{
            \begin{tabular}{l|c}
                \textbf{Num. Frames} & \textbf{Test Acc. (\%)} \\
                \Xhline{1.0pt}
                16 & 23.93 \\
                32 & 25.20 \\
                \rowcolor{gray!20} \textbf{64} & \textbf{28.50} \\
                128 & 27.18 \\
            \end{tabular}
        }
        \subcaption{
            \textbf{Number of Frames}: We vary the number of input frames and observe that 64 frames lead to the best accuracy. 
        }
        \label{ablation:frames}
    \end{minipage}
    \ 
    \begin{minipage}[t]{0.39\linewidth}
        \vspace{0pt}
        \centering
        \resizebox{\textwidth}{!}{
            \begin{tabular}{l|c|c}
                \textbf{Sampling Rate} & \textbf{Test Views} & \textbf{Test Acc. (\%)} \\
                \Xhline{1.0pt}
                32 & 8 & 22.64 \\
                64 & 4 & 25.89 \\
                128 & 2 & 26.30 \\
                \rowcolor{gray!20} \textbf{Uniform} & \textbf{1} & \textbf{28.50} \\
            \end{tabular}
        }
        \subcaption{
            \textbf{Sampling Rate}: We investigate different frame sampling rates (i.e., the interval between consecutive frames) and report that uniform sampling works the best. We use 64 frame inputs for these experiments.
        }
        \label{ablation:sampling_rate}
    \end{minipage}
    \
    \begin{minipage}[t]{0.3\linewidth}
        \vspace{0pt}
        \centering
        \resizebox{\textwidth}{!}{
            \begin{tabular}{l|c}
                \textbf{Data Fraction (\%)} & \textbf{Test Acc. (\%)} \\
                \Xhline{1.0pt}
                25 & 23.67 \\
                50 & 24.89 \\
                75 & 27.76 \\
                \rowcolor{gray!20} \textbf{Full} & \textbf{28.50} \\
            \end{tabular}
        }
        \subcaption{
            \textbf{Data Fraction}: We vary the amount of training data and observe that using full data leads to the best accuracy.
        }
        \label{ablation:ratio}
    \end{minipage}
    }  

    
    \caption{We ablate different design choices of the best performing VideoMamba model. The skill estimation performance is evaluated using the top-1 accuracy metric averaged across 20 skills.}
    \vspace{-0.1cm}
    \label{ablations}
    \vspace{-0.2cm}
\end{table*}

To conduct our human study, we used videos from the 2018–2019 NCAA Division I season. 
During the study, we did not provide any information about the league or the players to the subjects to avoid bias.
Furthermore, subjects were asked if they recognized any players before the study and were disqualified if they answered ``yes". To reduce the variance in human scores, we asked each subject to perform the study on 3 sets of different players. Additionally, for a comparison with a computer vision model, we included our best-performing VideoMamba~\cite{li2024videomamba} baseline, which we fine-tuned for three-level skill classification to match the setting of the human study.  Note that all players in the human study were sampled from the test set to make the comparison fair.

In Figure \ref{img:human_study}, we present our human subjects' results, which include data from $11$ subjects. We categorize the results based on the subjects' expertise level (i.e., novice, average, expert). For each group, we calculate the average performance and plot the maximum and minimum results. Based on the results, we first report that VideoMamba achieves \textbf{41\%} accuracy (compared to \textbf{33.3\%} random baseline). We also observe that human subjects in the novice group achieved \textbf{43\%}, barely outperforming VideoMamba. However, when considering human subjects with higher expertise, we observe that their accuracy is much higher, with the expert group averaging \textbf{66\%} with a top performance reaching \textbf{72\%}. These results suggest that human experts can solve this skill estimation task while computer vision models struggle.

\subsection{Generalization Analysis} 

In Table \ref{tab:ablation_breakdown}, we analyze how our best-performing VideoMamba model generalizes to 1) videos across different seasons, 2) videos with players of different gender, and 3) videos with different geographic locations. Our results suggest several interesting trends. First, we observe that the model is doing slightly better in videos from more recent seasons. This can be attributed to the fact that the more recent seasons have more videos. Next, we observe that the model is doing better in the videos of female players, which is somewhat surprising since our training videos contain $3\times$ more videos of male players. This result signifies that the model generalizes reasonably well between genders. Lastly, our results indicate that the model achieves the best accuracy in videos originating from North America and the worst accuracy in videos from Australia. We hypothesize that this is because we have $55\times$ more videos from North America than from Australia.

\subsection{Cross-Season \& Cross-League Generalization} 

Next, in Table~\ref{zero_shot}, we perform evaluations of VideoMamba for cross-season and cross-league generalization in previously unseen settings. Specifically, we define three evaluation scenarios: 1) testing on videos from previously unseen seasons, 2) testing on videos from previously unseen geographic locations (i.e., leagues), and 3) the combination of 1) and 2). Our results in Table~\ref{zero_shot} suggest that compared to the testing accuracy obtained on the in-domain data (i.e., previously seen season and league), the results on previously unseen leagues drop substantially ($>4\%$). We also observe that generalizing across leagues is more difficult than across seasons. 

\newcommand{\cmark}{\ding{51}}%
\newcommand{\xmark}{\ding{55}}%

\begin{table}[t]
    \small
    \centering
    \setlength\tabcolsep{15pt}
    \begin{tabular}{l l | c}
         \toprule
         \textbf{Season} & \textbf{League} & \textbf{Test Acc. (\%)}\\ 
         \hline
        Unseen & Unseen & 23.53 \\
        Unseen & Seen & 26.83 \\
        Seen & Unseen & 23.09 \\ 
        Seen & Seen & 28.50 \\

        \bottomrule
    \end{tabular}
    \caption{We conduct cross-season \& cross-league generalization evaluations of the best-performing VideoMamba model to test its performance on various previously unseen scenarios. We observe that the model performs substantially worse on the videos from previously unseen seasons and locations, signifying issues with out-of-domain generalization. 
    }
    \vspace{-0.3cm}
    \label{zero_shot}
\end{table}

\subsection{Ablation Studies}

In Table~\ref{ablations}, we use VideoMamba~\cite{li2024videomamba} to perform ablation studies on various model design choices, including 1) the number of input frames, 2) different frame sampling strategies, 3) the effect of the training data size, and 4) varying the clips included in the player highlight video.

\noindent \textbf{Number of Input Frames.} In Table~\ref{ablation:frames}, we investigate the model's performance with a different number of input frames. For these experiments, we use the uniform frame sampling strategy. Although the model can process 128 frames, we find that using 64 frames yields the best results.

\noindent \textbf{Frame Sampling Rate.} In Table~\ref{ablation:sampling_rate}, we test different frame sampling rates with a fixed input of 64 frames. We also vary the number of temporal test views to cover the whole video input during inference. We observe that uniform sampling leads to the highest accuracy.

\noindent \textbf{Training on the Subset of Data.} In Table~\ref{ablation:ratio}, we study the skill estimation performance as a function of the training data size. Our results indicate that using the full data leads to the best performance. 

\noindent\textbf{Varying the Selected Player Clips.} Lastly, we study how varying the clips in the player's highlight video affects the performance. To do this, we generated 5 different highlight videos for each player by randomly sampling 50 clips each time. We report the average accuracy of 28.44\%, with a variance of 1.30, which is comparable to our reported accuracy of 28.50\%. Thus, based on these results, we can conclude that the selection of clips does not dramatically impact the overall results.

\section{Conclusion}

We introduce \name, a diverse, large-scale basketball video dataset for fine-grained skill estimation. Our experiments with many recent state-of-the-art video recognition models revealed that all of these models struggle to achieve high skill estimation accuracy, thus highlighting the need for video models with fundamentally new recognition capabilities. We hope that our new dataset will inspire new ideas in fine-grained skill estimation and will lead to many practical applications, from promoting fair basketball scouting to developing personalized player development tools.

{
    \small
    \bibliographystyle{ieeenat_fullname}
    \bibliography{main}
}

\clearpage
\setcounter{page}{1}
\maketitlesupplementary

Our supplementary material consists of Additional Dataset Details (Section~\ref{sec:additional_dataset_details}), Additional Implementation Details (Section~\ref{sec:additional_implemetation_details}), Additional Human Evaluation Details (Section~\ref{sec:additional_evaluation_details}), and Additional Results Analysis (Section~\ref{sec:additional_results_analysis}).

\section{Additional Dataset Details}
\label{sec:additional_dataset_details}

In this section, we provide additional details of our \name~dataset.

\begin{figure*}[!t]
    \centering
    \includegraphics[width=1.95\columnwidth]{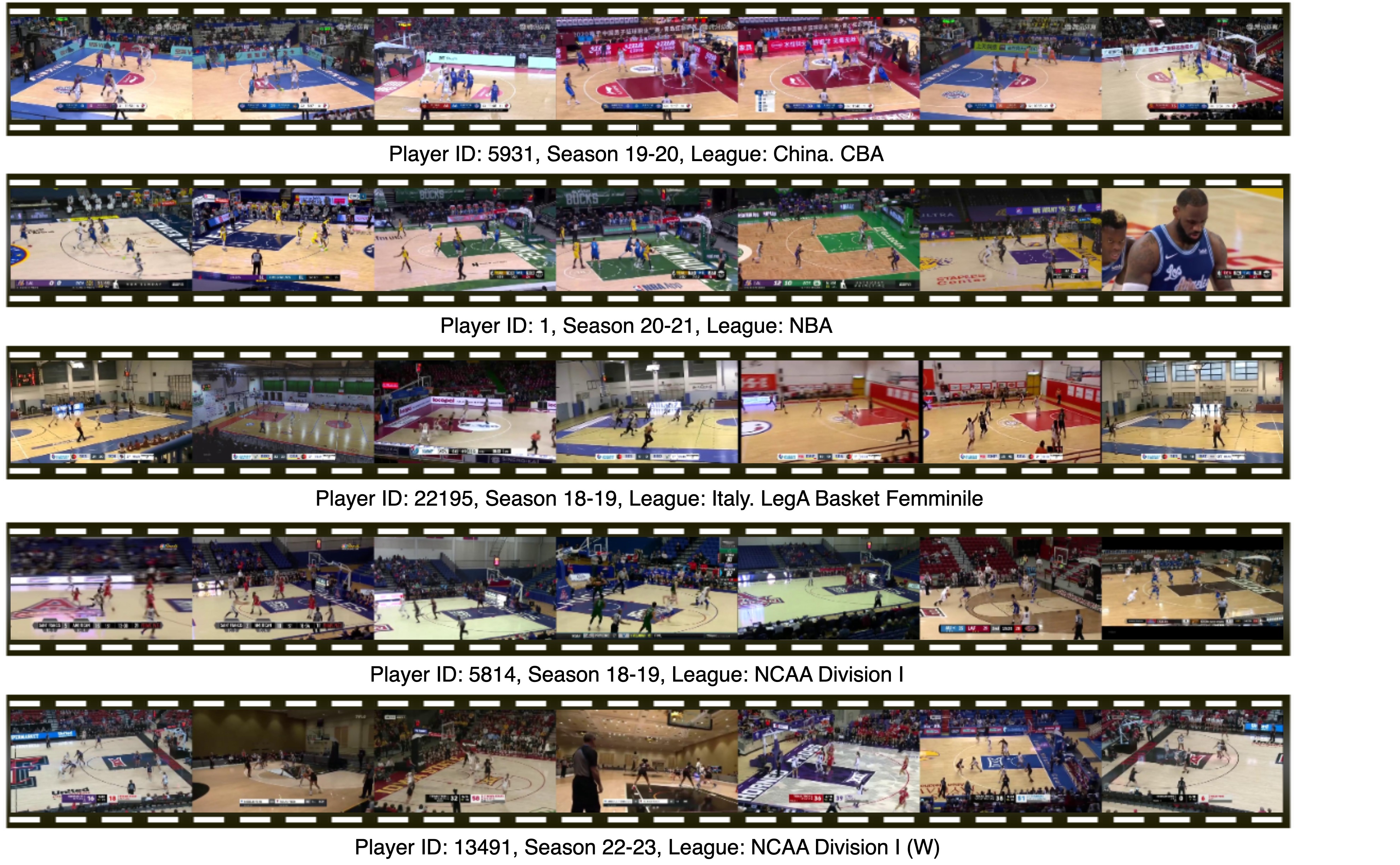}
    \caption{
    \textbf{Snapshots from the Player Highlight Videos.} Each row represents the video of a particular player. For these visualizations, we uniformly sample eight frames from the input video.
    } 
    \label{img:snapshots}
\end{figure*}

\noindent\textbf{Visualization of Player Highlight Videos.}  Figure \ref{img:snapshots} showcases snapshots from five randomly selected player highlight videos used as inputs to the skill estimation models. For each video, we visualize eight uniformly sampled frames from the full highlight video.

\noindent\textbf{Additional Details on Constructing Skill Labels.}  We also note that different leagues (e.g., college vs. professional) may exhibit skill-level differences. To account for these variations and avoid player comparisons across different leagues, we construct ground truth skill labels separately for each league. During model training/inference, the model then has to implicitly learn to predict skill levels that are specific to each basketball league.

\noindent\textbf{Dataset Characteristics Analysis.} Table \ref{tab:player_number_breakdown} presents the distribution of players by seasons, player genders, and geographic locations. These numbers highlight the richness and diversity of our proposed dataset.

\section{Additional Implementation Details}
\label{sec:additional_implemetation_details}

In this section, we provide additional details of model implementation. 

\noindent\textbf{Model Hyperparameters.} In Table \ref{tab:training_spec}, we provide additional details about the hyperparameters used for all of our tested models.

\begin{figure*}[!t]
    \centering
    \vspace{0cm}
    \includegraphics[width=1.95\columnwidth]{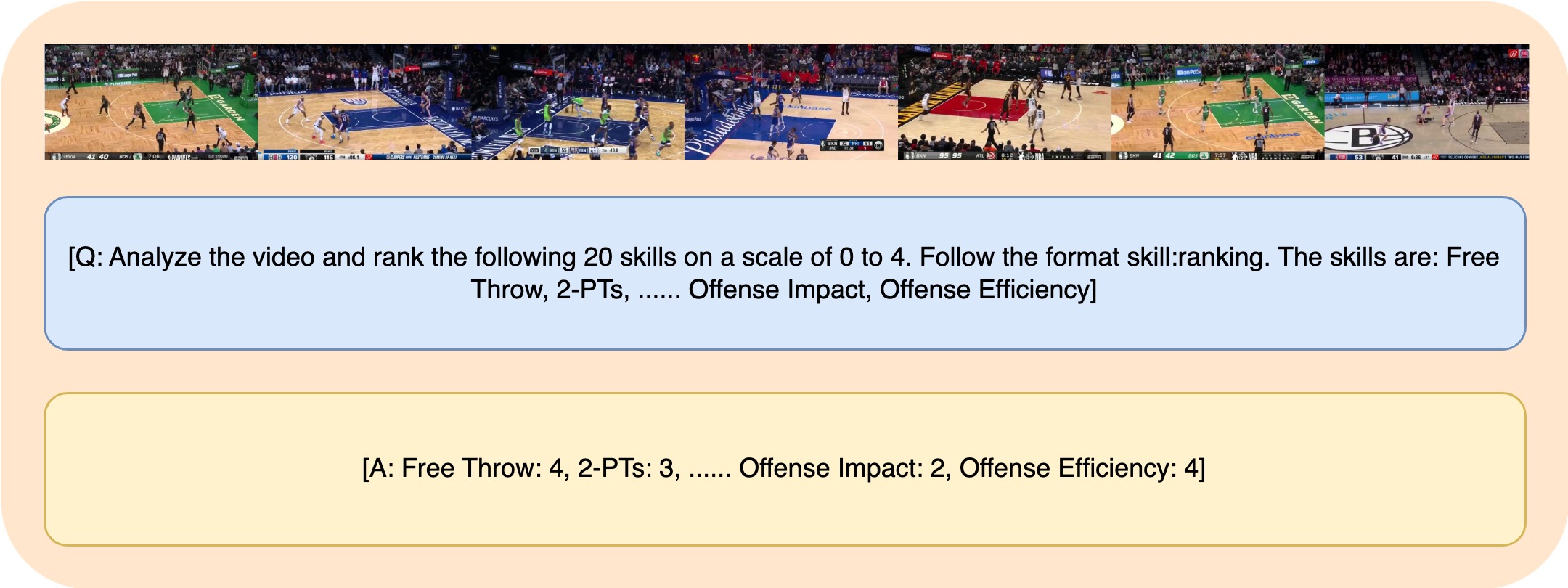}
    \caption{
    \textbf{A Sample Prompt Used to Train LLaVA-OneVision Model.} We convert our \name~dataset into a text-based instruction tuning format commonly used by modern VLMs. The model takes the instruction to assign a numerical category to each of the twenty basketball skills and then generates a textual answer. We then evaluate the model's performance using top-1 accuracy based on the generated outputs.
    } 
    \label{img:prompt_demo}
    \vspace{-0.3cm}
\end{figure*}

\noindent\textbf{LLaVA-OneVision Implementation.} To fine-tune LLaVA-OneVision, we reformat our \name~dataset into a text-based instruction-tuning format, commonly used by modern VLMs. Figure \ref{img:prompt_demo} provides a sample prompt used as input to the LLaVA-OneVision model. Specifically, we prompt the model to assign a numerical category to each of the twenty skills. We then evaluate the model's performance using top-1 accuracy based on the generated outputs.

\begin{table*}[b]
    \centering
    \renewcommand{\arraystretch}{1.25}
    \setlength\tabcolsep{5pt} 
    
    \begin{subtable}[t]{\linewidth}
        \centering
        \begin{tabular}{c | c c c c c c}
            \textbf{Season} & 17-18 & 18-19 & 19-20 & 20-21 & 21-22 & 22-23 \\
            \Xhline{1.0pt}
            \textbf{Num. Players} & 2360 & 3736 & 6047 & 5073 & 6879 & 8137 \\
        \end{tabular}
        \caption{Number of Players by Season}
        \label{tab:ablation_season}
    \end{subtable}
    \vspace{0.4cm}

    \begin{subtable}[t]{\linewidth}
        \centering
        \begin{tabular}{c | c c}
            \textbf{Gender} & Male & Female \\
            \Xhline{1.0pt}
            \textbf{Num. Players} & 24669 & 7563 \\
        \end{tabular}
        \caption{Number of Players by Gender}
        \label{tab:ablation_gender}
    \end{subtable}
    \vspace{0.4cm}

    \begin{subtable}[t]{\linewidth}
        \centering
        \begin{tabular}{c | c c c c}
            \textbf{Location} & N. America & Europe & Asia & Australia \\
            \Xhline{1.0pt}
            \textbf{Num. Players} & 22632 & 8177 & 1045 & 378 \\
        \end{tabular}
        \caption{Number of Players by Location}
        \label{tab:ablation_continent}
    \end{subtable}
    
    \caption{\textbf{Breakdown of Player Numbers.} We present the detailed player numbers of \name~by (1) seasons, (2) gender, and (3) geographic locations, highlighting the diversity of our dataset.}
    \label{tab:player_number_breakdown}
\end{table*}

\begin{table*}[!t]
    \centering
    \setlength\tabcolsep{3pt}
    \renewcommand{\arraystretch}{1.15}
    \begin{tabular}{l l l l l l l l l l}
        \toprule
        \textbf{Attribute} & \textbf{LLaVA-OV} & \textbf{MeMViT} & \textbf{SigLIP} & \textbf{VideoMAE2} & \textbf{X-CLIP} & \textbf{TSF} & \textbf{UMT} & \textbf{IV2} & \textbf{VideoMamba}\\
        \hline
        \textbf{LR} & 1e-5 & 5e-2 & 1e-5 & 7e-4 & 1e-6 & 1e-2 & 7e-3 & 6e-5 & 3e-4 \\
        \textbf{Epoch}         & 1    & 20   & 20   & 20    & 20   & 20   & 20   & 20    & 20    \\
        \textbf{Warmup Epochs} & 0 & 0 & 0 & 2 & 0 & 0 & 2 & 2 & 2 \\
        \textbf{Batch Size}    & 1    & 16   & 8    & 1     & 8    & 16   & 2    & 4     & 4     \\
        \textbf{Optimizer}     & AdamW & SGD  & Adamw & Adamw & Adamw & SGD & Adamw & Adamw & AdamW \\
        \textbf{Drop Path Rate}& 0    & 0.5  & 0    & 0.1   & 0    & 0.5  & 0.1  & 0.3   & 0.5   \\
        \bottomrule
    \end{tabular}%
    \caption{\textbf{Summary of Hyperparameters for Different Models}.
    \textbf{TSF}: TimeSformer, \textbf{UMT}: UnmaskedTeacher, \textbf{IV2}: InterVideo2
    }
    \label{tab:training_spec}
    \vspace{-0.2cm}
\end{table*}

\section{Human Evaluation Details} 
\label{sec:additional_evaluation_details}

Our user study is conducted online. To improve the reliability of the results and incentivize users, we used a two-level compensation scheme. Specifically, participants would receive a base compensation for completing each session with a bonus if their performance exceeded a certain accuracy threshold (i.e., 60\%).

To optimize the study design, we conducted a preliminary test study with three participants to determine the appropriate number of player videos to review and the skills to evaluate. Based on feedback from this initial study, we finalized the study design to include the assessment of five players across five skills, each categorized into three levels of skills. We observed that evaluating a single-player video required approximately 10-12 minutes, and participants demonstrated consistent scoring accuracy for skills within the same coarse category. We also observed a noticeable decline in participant performance after one hour of video review, hence the motivation for the study design, where a single session could be completed within one hour.

To recruit participants for the study, we distributed advertisements in computer science department channels and online basketball group chats. Participants were asked about their basketball experience, including the number of years they spent watching or playing the sport. Based on their responses, all participants were categorized into novice, average, or expert groups. 

\section{Additional Results Analysis}
\label{sec:additional_results_analysis}

In Table \ref{tab:skill_breakdown}, we provide a detailed breakdown of how the best-performing VideoMamba model generalizes across different skill categories included in our \name~dataset. Our results suggest that the average accuracy for coarse skill categories such as shooting, rebounding, defense, and playmaking are relatively similar. 
However, we also observe that offensive skills generally exhibit higher accuracy. We believe that the differences in accuracy may be attributed to the fact that many of the video clips in the player highlight videos are offensive plays. In contrast, defensive players are rarer and therefore more challenging for the model to learn to assess.

\begin{table*}[h]
    \centering
    \setlength\tabcolsep{18pt}
    \begin{tabular}{l r}
    \toprule
    \textbf{Skill} & \textbf{Test Acc. (\%)} \\
    \midrule
    \textbf{Shooting}  &  \\
    Free Throw         & 27.76 \\
    2-PTs              & 27.62 \\
    3-PTs              & 24.14 \\
    Contested-shots    & 26.62 \\
    Overall Shooting   & 26.19 \\
    \rowcolor{gray!10} Average  & 26.47 \\
    \midrule
    \textbf{Rebounding}&  \\
    Rebounds           & 27.11 \\
    Defensive          & 26.85 \\
    Offensive          & 25.89 \\
    \rowcolor{gray!10} Average  & 26.62 \\
    \midrule
    \textbf{Defense}   &  \\
    Steals             & 24.76 \\
    Fouls              & 25.79 \\
    Points-allowed     & 27.76 \\
    Defensive Consistency & 31.20 \\
    \rowcolor{gray!10} Average  & 27.38 \\
    \midrule
    \textbf{Playmaking}&  \\
    Assists            & 25.89 \\
    Passing Accuracy   & 24.69 \\
    Turnovers          & 25.59 \\
    \rowcolor{gray!10} Average  & 25.39 \\
    \midrule
    \textbf{Offense}   & \\
    Contribution       & 35.09 \\
    Offensive Consistency & 34.34 \\
    Teamwork           & 30.17 \\
    Impact             & 37.24 \\
    Efficiency         & 36.14 \\
    \rowcolor{gray!10} Average  & 34.60 \\
    \bottomrule
    \end{tabular}
    \caption{\textbf{Accuracy Breakdown Across Skills.} We take our best-performing VideoMamba and breakdown its accuracy on the twenty fine-grained. We observe that the average accuracy for the four coarse skill categories of shooting, rebounding, defense, and playmaking are relatively similar. Additionally, we observe that the accuracy for offensive skills is slightly higher, suggesting that these skills might be easier to predict.}
    \label{tab:skill_breakdown}
    \vspace{-0.2cm}
\end{table*}

\end{document}